\documentclass{article}

\usepackage{microtype}
\usepackage{graphicx}
\usepackage{subfigure}
\usepackage{booktabs} 

\usepackage{amsfonts}
\usepackage[table, svgnames, dvipsnames]{xcolor}
\usepackage{enumitem}
\usepackage{mathrsfs}
\usepackage{microtype}
\usepackage{booktabs} 
\usepackage{multirow}
\usepackage[textsize=tiny]{todonotes}
\usepackage{tcolorbox}
\usepackage{soul}
\usepackage{wrapfig}
\usepackage{tikz}
\usepackage{graphicx}
\usepackage{subfigure}
\usepackage[square,sort,comma,numbers]{natbib}
\usepackage{svg}

\usepackage{hyperref}
\usepackage{soul} 

\usepackage[accepted]{icml_fmwild}

\usepackage{amsmath}
\usepackage{amssymb}
\usepackage{mathtools}
\usepackage{amsthm}

\usepackage[capitalize,noabbrev]{cleveref}

\theoremstyle{plain}

\theoremstyle{definition}

\theoremstyle{remark}

\usepackage[textsize=tiny]{todonotes}

\icmltitlerunning{FLUID-LLM: Learning Computational Fluid Dynamics with Spatiotemporal-aware Large Language Models}

\begin{document}

\twocolumn[
\icmltitle{FLUID-LLM: Learning Computational Fluid Dynamics with \\ Spatiotemporal-aware Large Language Models}



\icmlsetsymbol{equal}{*}

\begin{icmlauthorlist}
\icmlauthor{Max Zhu}{equal,aff1}
\icmlauthor{Adrián Bazaga}{equal,aff1}
\icmlauthor{Pietro Liò}{aff1}
\end{icmlauthorlist}

\icmlaffiliation{aff1}{University of Cambridge, Cambridge, United Kingdom}

\icmlcorrespondingauthor{Max Zhu}{mz406@cam.ac.uk}
\icmlcorrespondingauthor{Adrián Bazaga}{ar989@cam.ac.uk}

\icmlkeywords{Computational Fluid Dynamics, Learning Fluid Dynamics, Language Models, Deep Learning}

\vskip 0.3in
]



\printAffiliationsAndNotice{\icmlEqualContribution} 

\begin{abstract}

Learning computational fluid dynamics (CFD) traditionally relies on computationally intensive simulations of the Navier-Stokes equations. Recently, large language models (LLMs) have shown remarkable pattern recognition and reasoning abilities in natural language processing (NLP) and computer vision (CV). However, these models struggle with the complex geometries inherent in fluid dynamics. We introduce FLUID-LLM, a novel framework combining pre-trained LLMs with spatiotemporal-aware encoding to predict unsteady fluid dynamics. Our approach leverages the temporal autoregressive abilities of LLMs alongside spatial-aware layers, bridging the gap between previous CFD prediction methods. Evaluations on standard benchmarks reveal significant performance improvements across various fluid datasets. Our results demonstrate that FLUID-LLM effectively integrates spatiotemporal information into pre-trained LLMs, enhancing CFD task performance.

\end{abstract}

\section{Introduction}

Computational fluid dynamics (CFD) is important in many areas of science and engineering. It is particularly relevant in the design of air vehicles \cite{Blake2022} and civil infrastructures \cite{Zheng2010}, modeling biomedical flows and supporting medical device design \cite{Doost2016} or understanding environmental phenomena \cite{Kim1999}. However, solving the Navier-Stokes equations governing fluid mechanics is a fundamental challenge in computational physics, requiring intensive numerical simulations and extensive computational resources. Traditional approaches to computational fluid dynamics (CFD) such as finite volume or finite elements \cite{Bassi1997}, often demand weeks of computation for complex problems and necessitate expert configuration of numerical solvers. 

Recently, advancements in machine learning \cite{Brunton2020,Vinuesa2022,Lino2023}, have shown promise in simulating fluid dynamics with reduced computational overhead \cite{Lino2022,Hu2023}. Concurrently, large language models (LLMs) have demonstrated significant capabilities in pattern recognition and reasoning across complex sequences in multiple modalities, such as text \cite{brown2020language,ouyang2022training} or image \cite{Radford2021LearningTV,koh2023generating}. These models have achieved notable success in generalizing to tasks not directly related to language with minimal task-specific modifications, leading to their application in diverse domains, including time series forecasting \cite{zhang2024large, hota2024evaluating, liu2024lstprompt, jin2024time, xue2023promptcast}. However, directly applying time-series LLM techniques to fluid dynamics poses unique challenges due to high dimensional spatial and temporal dependencies inherent in fluid flow data \cite{Eagle, li2023scalable} where each timestep contains several thousands of data points. 

\textbf{Contributions.} We introduce FLUID-LLM, a novel framework that integrates pre-trained LLMs with spatiotemporal-aware encoders and decoders to predict unsteady fluid dynamics. By leveraging the autoregressive abilities of LLMs, FLUID-LLM bridges the gap between LLM-based and GNN-based approaches for CFD prediction. This integration allows for the effective utilization of spatiotemporal information in pre-trained LLMs, enhancing their performance on fluid dynamics tasks. We evaluate FLUID-LLM on standard CFD benchmarks, demonstrating significant improvements in performance across fluid datasets. Furthermore, the use of language pretrained LLMs improves performance and allows for in-context learning abilities. We envision our work to offer a promising direction for efficiently addressing the computational challenges of fluid dynamics simulations.

\begin{figure*}[t]
    \centering
\includegraphics[width=0.97\textwidth]{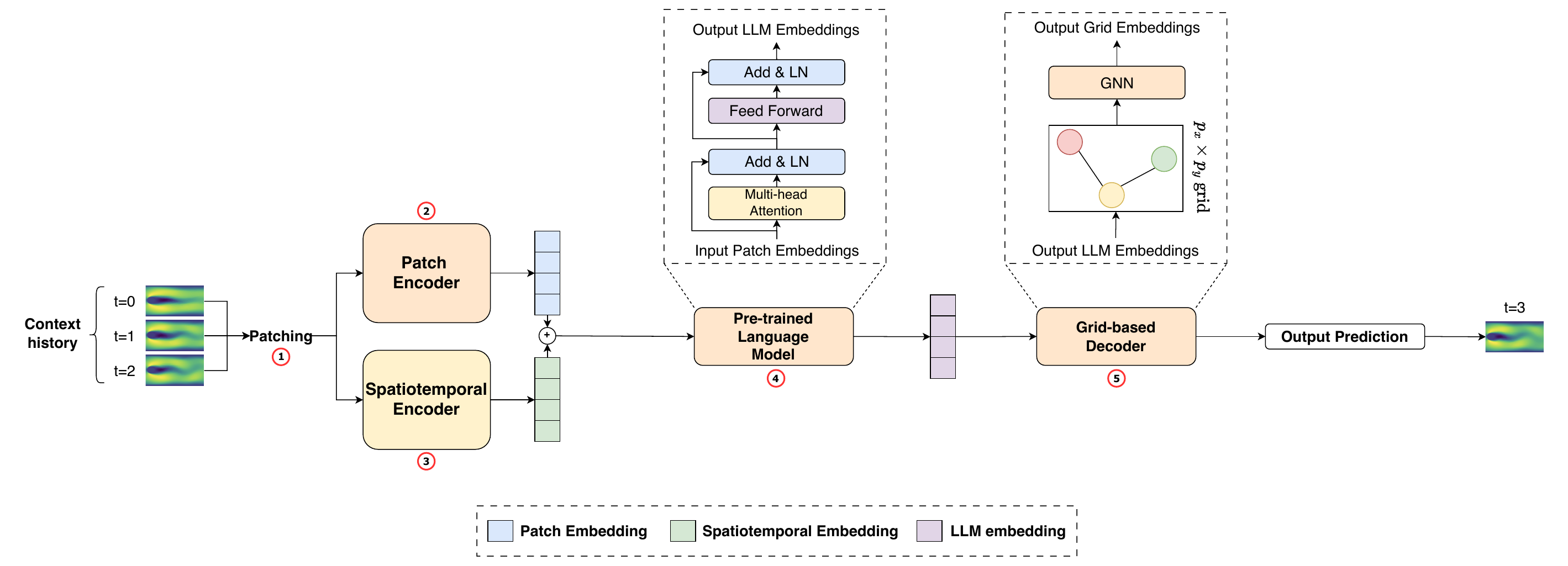}
    \caption{High-level overview of the FLUID-LLM framework. Given an input context history of fluid states, we first tokenize (1) and embed it via patch encoding (2) along with a spatiotemporal embedding (3). These combined embeddings are then given as input to a fine-tuned language model (LLM). Then, the LLM produces updated embeddings by forwarding them through its internal layers (4). These are projected into a grid, updating each state embedding with a GNN to allow for information propagation between states in the grid (5). Finally, the network predicts the differences between the previous and next state to generate the subsequent fluid state.}
    \label{fig:FLUIDLLM}
\end{figure*}

\section{Methodology}

Our model architecture is depicted in Figure~\ref{fig:FLUIDLLM}. At the core of our model is a fine-tuned LLM that makes predictions in a autoregressive fashion, taking in a history of previous states and with spatiotemporal-aware embeddings before predicting the next states. Specifically, our problem consists of a 2-dimensional fluid flow simulation, with $x$ and $y$ describing velocity ($V_x$ and $V_y$) and pressure ($P$). Our task is, given a sequence of simulation states $s_0, ..., s_T$ up to timestep $T$, forecast the future velocity and pressure components.


\subsection{Model Structure}

\textbf{Model Inputs.} In order to apply LLMs to fluid flow data, we insert a patch-based encoder \cite{dosovitskiy2021an} that transforms continuous spatial fluid flow to the feature space of the LLM. Firstly, each timestep $t$ of the dataset $s_t$ is converted to a 2D regular grid, where the $x$ and $y$ components of velocity ($V_x$ and $V_y$) and pressure ($P$) are stacked as channels (along with other features such as density if present). Let the 2D grid have dimensions $p_x \times p_y$, then each state at step $t$ has shape $s_t \in \mathbb{R}^{p_x \times p_y \times 3}$. Boundary conditions are set by fixing the corresponding pixel in all channels to a fixed value. To handle fluid simulations with irregular or dynamic meshes while being independent of the mesh/measurement generation process, we project the original fluid mesh onto a regular grid via linear interpolation. 


\textbf{Input Embedding.} Given a sequence of $\tau$ previous states $s_1, ... s_\tau$ (where $\tau = 1$ for the initial state), a flattened input sequence of embeddings is generated by patching. Specifically, each state is split into $N = p_x \cdot p_y / 16^2$ patches of size $(16 \times 16)$, $s_t'\in \mathbb{R}^{N \times 16 \times 16 \times 3}$ for each timestep $t$. The patches are ordered sequentially in the $x$ or $y$ direction depending on the direction of the fluid flow if the simulation is directional. Next, to produce patch embeddings, a patch encoder $f_{\text{enc}}: \mathbb{R}^{16 \times 16 \times 3} \to \mathbb{R}^{d_\text{llm}}$, encodes each patch independently to generate embeddings $e^\text{enc}_t = f_{\text{enc}}(s_t')\in \mathbb{R}^{N \times d_{\text{llm}}}$, where $d_{\text{llm}}$ corresponds to the dimensionality of the LLM feature space. Finally, the embeddings for each timestep are concatenated to form a single sequence of patch embeddings $e^\text{enc} \in \mathbb{R}^{\tau \cdot N \times d_\text{LLM}}$. 

\textbf{Spatiotemporal-Aware Embedding.} To augment the patch embeddings with location and time information, an spatiotemporal encoding is added onto the patch embeddings that allows the subsequent LLM to identify the location and time dependencies for each patch. In particular, for a patch with spatial and time coordinates $(x, y)$ and $t$, respectively, a learned spatiotemporal embedding maps the $x$ and $y$ locations, as well as the timestep $t$, into three vectors of dimensionality $d_\text{llm} / 3$. These embeddings are then concatenated to become the patch spatiotemporal embedding $e^\text{spatiotemporal} \in \mathbb{R}^{d_\text{llm}}$, which are added to the initial patch embedding to generate the spatiotemporal-aware patch embedding $\tilde{e}^\text{enc} = e^\text{enc} + e^\text{spatiotemporal}$.

\textbf{Language Model Embedding.} Before being input into the LLM, an additional copy of the first timestep embeddings is concatenated back onto the full embeddings to ensure the LLM can see the full initial state before making predictions, 
$\tilde{e}^\text{enc}_\text{all} = \tilde{e}^\text{enc}_{t=1} || \tilde{e}^\text{enc} \in \mathbb{R}^{(\tau+1) \cdot N \times d_\text{LLM}}$. 
The LLM, $f_\text{LLM}: \mathbb{R}^{d_\text{LLM}} \to \mathbb{R}^{d_\text{LLM}}$, generates output embeddings as $e^\text{out} = f_\text{LLM}(\tilde{e}^\text{enc}_\text{all}) \in \mathbb{R}^{(\tau + 1) \cdot N \times d_\text{LLM}}$ . The first $N$ vectors of $e^\text{out}$ are discarded and the remaining $\tau \cdot N$ outputs are split into $\tau$ output embeddings, $e^\text{out}_t \in \mathbb{R}^{N \times d_\text{LLM}}$ for each timestep $t \in \{1, \ldots, \tau \}$, which are passed to a grid-based decoder to predict the next state separately.

\textbf{Grid-based Decoding.} The decoder $f_\text{dec}: \mathbb{R}^{N \times d_\text{LLM}} \to \mathbb{R}^{p_x \times p_y \times 3}$ predicts the difference between the current state $s_t$ and the next state $s_{t+1}$. Therefore, the next state can be predicted from the previous state as $\hat{s}_{t+1} = s_t + f_\text{dec}(e^\text{out}_t)$. Specifically, the decoder consists of a MLP, $f_\text{proj}: \mathbb{R}^{d_\text{LLM}} \to \mathbb{R}^{16 \times 16 \times d_\text{GNN}}$ that projects each LLM output token onto a $16 \times 16$ patch representation, which are concatenated together into a single $(p_x \times p_y$) regular grid and decoded with a Graph Attention Network \cite{GAT, GatV2Conv}, $f_\text{GNN}: \mathbb{R}^{p_x \times p_y \times d_\text{LLM}} \to \mathbb{R}^{p_x \times p_y \times 3}$. This GNN predicts the difference between $s_t$ and $s_{t+1}$. 



\textbf{Training setup.} Training is performed autoregressively on every timestep $t$ simultaneously. The encoder $f_\text{enc}$ and decoder $f_\text{dec}$ are initialized randomly, while the LLM $f_\text{llm}$ is initialized from a pretrained checkpoint and fine-tuned using DoRA \cite{dora}. The entire architecture is trained end-to-end. To achieve autoregressive inference, each next state predicted is added to the history of states. Note that while the next timestep is predicted from the previous timestep, the next patch does not depend on the previous patch but on the last timestep $N$ patches in the past. The advantage of this setup is the LLM can make predictions for the next state (or equivalently $N$ patches) in parallel, greatly improving inference efficiency. Further details of our model and design choices are given in Appendix \ref{appendix:Implementation}. 

\section{Experiments and Results}

\begin{table*}[ht]
    \centering
    \caption{Prediction RMSE ($\downarrow$) on the Cylinder and Airflow datasets for our model and baselines after a different number of steps. Best model highlighted in \textbf{bold}. Divergent predictions are given as \textit{NA}.}
    \label{tab:main_results}
    \resizebox{0.95\textwidth}{!}{
        \begin{tabular}{l|cccc|cccc}
        \hline
        \multirow{2}{*}{\textbf{Method}} & \multicolumn{4}{c|}{\textbf{Cylinder}} & \multicolumn{4}{c}{\textbf{Airflow}} \\
        \cline{2-9}
        & \textbf{N=1} & \textbf{N=50} & \textbf{N=100} & \textbf{N=150} & \textbf{N=1} & \textbf{N=50} & \textbf{N=100} & \textbf{N=150} \\
        \hline
        DilResNet & 0.004 & 0.084 & 2.248 & \textit{N/A} & 0.011 & 0.266 & 8.711 & \textit{N/A} \\
        GAT2Conv & 0.004 & 0.070 & 0.117 & 0.127 & 0.011 & 0.776 & 2.796 & \textit{N/A} \\
        MeshGraphNets & 0.003 & 0.038 & 0.058 & 0.076 & 0.011 & 0.224 & 0.550 & 0.958 \\
        Random-OPT125m & 0.029 & 0.040 & 0.072 & 0.105 & 0.014 & 0.284 & 0.609 & 0.822 \\
        \hline
        FLUID-OPT125m (ours) & 0.002 & 0.031 & 0.062 & 0.102 & 0.011 & 0.201 & 0.411 & 0.716 \\
        FLUID-OPT2.7b (ours) & \textbf{0.002} & \textbf{0.023} & \textbf{0.041} & \textbf{0.059} & \textbf{0.007} & \textbf{0.118} & \textbf{0.259} & \textbf{0.457} \\
        \hline
        \end{tabular}
    }
\end{table*}

\textbf{Datasets.} Evaluations are made on two fluid datasets from MeshGraphNets: a simpler Cylinder dataset consisting of incompressible fluid flows around a cylinder in a tube, and a more complex Airflow dataset consisting of compressible transonic airflows over different wing configurations. Details of dataset processing are given in Appendix \ref{appendix:Dataset}. 

\textbf{Setup and evaluation.} Two LLM models are evaluated: OPT-125m and OPT-2.7b, named as \textbf{FLUID-OPT125m} and \textbf{FLUID-OPT2.7b}, respectively. Both models use the same encoder and decoder setup. We compare FLUID-LLM against recent machine learning methods for 2D fluid prediction, \textbf{MeshGraphNets} \cite{MGN}, a message-passing graph neural network that operates on the simulation mesh and \textbf{DilResNet} \cite{dilresnet}, a convolutional neural network (CNN) based method that operates on a square grid designed for complex turbulent flows. Also, we compare against a \textbf{GAT2Conv} method \cite{GatV2Conv}, which our decoder $f_\text{GNN}$ also uses, in order to ablate how much the decoder contributes to our overall model performance. Lastly, we try a LLM with randomly initialized weights instead of a pretrained checkpoint (\textbf{Random-OPT125m}) to assess the impact of LLM pre-training on task performance. Details of baselines are given in Appendix \ref{appendix:Baseline}. All models are evaluated using the root mean squared error over the N-step prediction horizon (N-RMSE) as defined in \citet{Eagle}. Separate models are trained for each dataset on the train split and evaluated on the test split. 

\subsection{Fluid Prediction Performance}

Table \ref{tab:main_results} displays the prediction RMSE for the Cylinder and Airfoil datasets at prediction horizons of 1, 50, 100, and 150 steps. FLUID-OPT2.7b consistently outperformed all other models across all horizons. Specifically, scaling the FLUID models from 125 million to 2.7 billion parameters resulted in a substantial RMSE reduction of approximately 42\% at the 150-step horizon for the Cylinder dataset. This demonstrates the significant impact of increasing the LLM size while keeping the encoder and decoder architectures unchanged. Additionally, a comparison between FLUID-OPT125m and Random-OPT125m highlights the accuracy improvements gained from language pretraining, particularly for shorter sequences. On the more complex Airfoil dataset, the performance differences between models are even more pronounced. FLUID-OPT2.7b achieved a 48\% lower RMSE at 150 steps compared to the best baseline, MeshGraphNets. This further underscores the superior performance of FLUID-OPT2.7b in handling longer prediction horizons and more complex datasets.

Figures \ref{fig:CylinderPlots} and \ref{fig:AirfoilPlots} illustrate model predictions for specific samples from the Cylinder and Airfoil datasets, respectively. DilResNet and GAT2Conv exhibited unstable predictions over longer horizons, particularly for the Airfoil dataset. Both FLUID-OPT2.7b and MeshGraphNets provided reasonable long-term predictions for the Cylinder dataset. However, only FLUID-OPT2.7b maintained accurate long-term predictions for the Airfoil dataset. Furthermore, while the smaller FLUID-OPT125m model produced blurry predictions over extended horizons, FLUID-OPT2.7b’s predictions remained clear and precise.



\subsection{Evaluation of In-context Learning Performance}

Our setup successfully demonstrates in-context learning, defined as enhancing prediction accuracy for the next state $s_{t+1}$ using a history of previous states $s_{1, ..., t}$. We showcase two types: 1) improved generalization on unseen parameter spaces for fixed tasks, and 2) few-shot learning with different tasks identified by the context history.

\begin{figure}
    \centering
    \includegraphics[width=\linewidth]{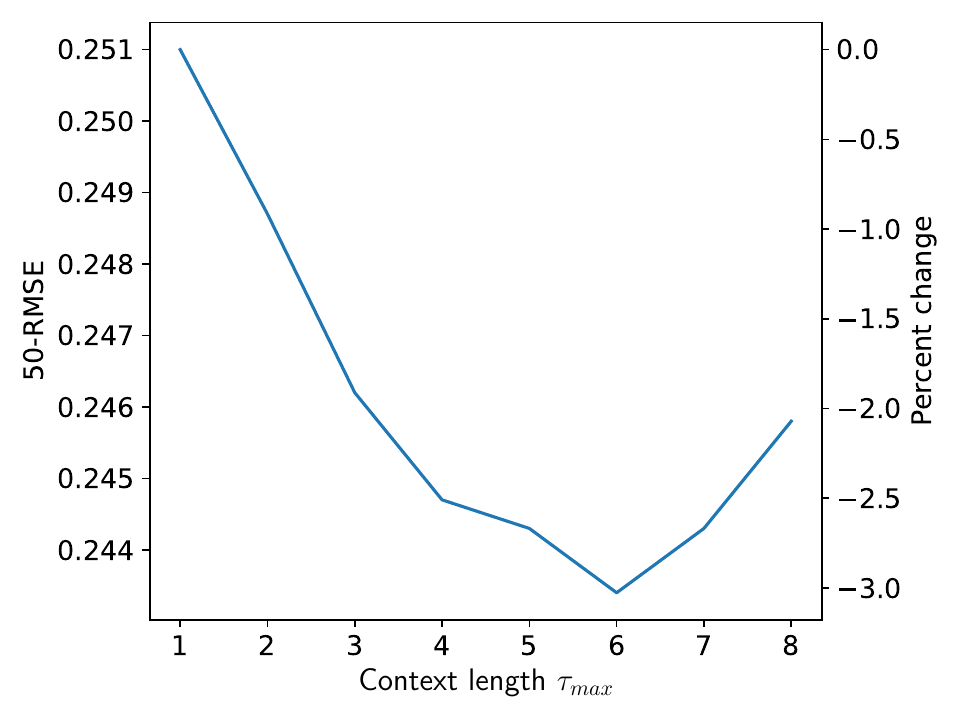}
    \caption{Predicted RMSE after 50 steps on the Airfoil dataset, where the model is given $\tau_\text{max}$ context states.}
    \label{fig:icl}
\end{figure}

For generalization, the context history is used to enhance performance on a fixed task. This is illustrated with the Airfoil test dataset, where simulation rules are constant, but the test parameter space differs from training. The training set includes Mach numbers from $[0.2, 0.7]$ and angles of attack (AOA) from $[-25^{\circ}, 25^{\circ}]$, while the test set uses Mach numbers from $[0.7, 0.9]$ and AOAs from $[-35^{\circ}, 35^{\circ}]$. Using the same FLUID-OPT2.7b model, we initialize it with a history of varying $\tau_\text{init}$ previous states and set the maximal LLM context length to $\tau_\text{max} = \tau_\text{init}$. The RMSE after 50 prediction steps is measured, with a constant starting prediction state for consistency. Figure~\ref{fig:icl} shows a 3\% error reduction by increasing $\tau_\text{init}$ from 1 to 6. This small reduction is due to the setup not strictly requiring in-context learning and the model not being explicitly trained for it. This property is likely retained from language pre-training and autoregressive fine-tuning.

\subsection{Few-shot Learning on Simulated Wave Evolution}

We demonstrate that our model is capable of learning simulated wave evolution by using few-shot learning. To evaluate this behavior, we generated a dataset of nonlinear 2D wave evolution governed by the following PDE:

\begin{equation} \label{eq:synth}
    \frac{d^2u(\mathbf{x}, t)}{dt^2} = c \cdot \text{erf} \left[\frac{0.1}{c} \cdot (\nabla^2 (\mathbf{x}, t) - \frac{d(\mathbf{x}, t)}{dt})\right]
\end{equation}

where $u(\mathbf{x}, t)$ is a scalar field, $\text{erf}$ is the error function, and $c$ is a scalar different for each sample, $c \sim \mathcal{U}(0, 5)$. Samples are generated by drawing $c$ from a uniform distribution, along with random boundary and initial conditions, and simulating 20 timesteps using an ODE solver \cite{NODE_solver}. Thus, the model must adapt to different values of $c$ from its context history. Further details are in Appendix \ref{Appendix:Synthetic}.

\begin{figure}
    \centering
    \includegraphics[width=\linewidth]{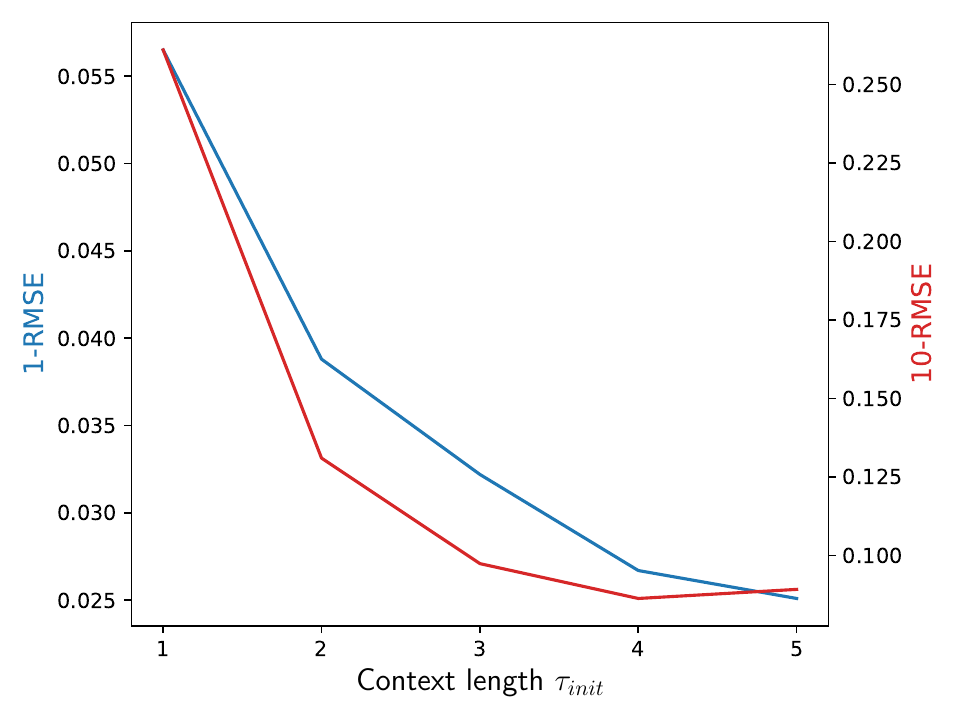}
    \caption{Predicted RMSE after 1 and 10 steps on the synthetic dataset, where the model is given $\tau_\text{init}$ initial context states.}
    \label{fig:synthetic}
\end{figure}

We trained a new FLUID-OPT125m model with the same autoregressive setup but with a shorter context length $\tau_\text{max} = 5$. To test the model, given an initial sequence of $\tau_\text{init}$ context observations, it predicts the next 10 steps. Figure~\ref{fig:synthetic} shows 1-RMSE and 10-RMSE decrease as $\tau_\text{init}$ increases, demonstrating the model learns the given equations of motion in context. Figure~\ref{fig:synth_preds} illustrates the model predictions. Details are given in Appendix \ref{Appendix:Synthetic}.

\section{Conclusions}

In this paper, we presented FLUID-LLM, a novel framework that harnesses pre-trained LLMs and integrates them with spatiotemporal encodings to predict fluid dynamics. FLUID-LLM bridges the gap between conventional CFD prediction techniques and modern machine learning approaches by enhancing the autoregressive capabilities of LLMs with spatiotemporal awareness. Our comprehensive evaluations demonstrate that FLUID-LLM enables the adaptation of LLMs into strong models for fluid dynamics. 


\bibliography{references}
\bibliographystyle{icml2024}

\newpage
\appendix
\onecolumn
\renewcommand{\thefigure}{A\arabic{figure}} 

\section{Implementation Details} \label{appendix:Implementation}

This section describes the implementation details of our model. For both OURS-125m and OURS-2.7b, the encoder $f_\text{enc}$ and decoder $f_\text{dec}$ parameters, remained the same, the only difference was the pre-trained OPT LLM was changed. The encoder $f_\text{enc}$ consists of a 2 layer MLP with 512 hidden layers with a LeakyReLU activation. Similarly, the decoder projection layer $f_\text{proj}$ is a MLP with the same 512 hidden layers and activation. The decoder GNN, $f_\text{GNN}$ is a 3-layer GNN using GATV2Conv layers with an input dimension of 32, hidden dimension 48, and output dimension of 3. This decoder GNN operates on a square graph, or between adjacent pixels of the output array. Since the GNN has 3 layers giving 3 pixels of message passing distance, and the patches are 16x16, the GNN can only modify small-scale features and long range features must be learned from the LLM. 

The LLMs are pre-trained models OPT-125m and OPT-2.7b checkpoints from Huggingface. Fine-tuning is performed using DORA low-rank adaptation on all layers with $r=16$, $\alpha=64$, and $\text{lora\_dropout} =0.1$ for both models. This allows around 0.4\% of the OPT-125m parameters and 0.2\% of the OPT-2.7b parameters to be trainable, meaning most of the model was fixed. Flash-Attention 2 \cite{flashattention2} was used to improve training and inference efficiency. 

The entire model was trained end-to-end with the ADAMW optimizer \cite{adamw} with an initial learning rate of 1e-3, decaying every 50 epochs by a factor of 0.75, and weight decay of 1e-5. The training loss is a weighted combination of mean squared error and mean absolute error, $L = \sum_t MSE(s_t, \hat{s}_t) + 0.01 \cdot MAE(s_t, \hat{s}_t)$. Following MeshGraphNet, 10 times higher weight is placed on velocity losses compared to pressure losses since pressure and velocity have different importance. For OPT-125m, we train for 180 epochs, while for OPT-2.67b, we train for 500 epochs. We found that training the smaller model for longer gave no improvements, while the larger model continued to improve with longer training duration. The sequence length during training was $\tau_\text{max}=10$ for FLUID-OPT125m and $\tau_\text{max}=8$ for FLUID-OPT2.7b, which is slightly reduced to speed up training and inference. 

FLUID-LLMs operate with patching on an interpolated grid if the original dataset is an irregular mesh. This is in contrast to GNN based methods such as MeshGraphNets which incorporate meshing as a part of the model. We argue this is unnecessary. Firstly, a GNN only allows message-passing between neighbors of a graph within each graph within each layer. While this may be beneficial by allowing more compute to be used on regions of high complexity, this restricts the distance information can propagate, whereas a transformer can attend to any point in its context and dynamically allocate compute to important regions. Secondly, while meshes are useful for numerical simulations, real-world fluids are continuous and do not exist on a mesh. 

\section{Dataset Processing} \label{appendix:Dataset}
Since the Cylinder and Airfoil datasets are irregular graphs, we project them onto regular grids using linear interpolation independently for each velocity component and pressure. The cylinder dataset contains around 1880 nodes in a rectangular domain and is projected onto a regular 240x64 grid giving 15x4 patches. The airfoil dataset contains around 5200 nodes on a circular domain, with large empty regions so we crop out a smaller central region between $-0.5<x<2$ and $0.75<x<0.75$ where the wing is and most interactions occur. This region is projected to a 240x144 grid giving 13x7 patches. This same cropping is done with all the baselines by discarding nodes outside this region, giving around 3680 nodes used for training and prediction.

Finally, since Fluid-OPT and DilResNet make predictions on regular grids, while MeshGraphNets and GAT2Conv make predictions on the irregular mesh, we experimented with how to directly compare results. We found that interpolating model predictions on the irregular mesh and onto regular grids (in the same way as states are interpolated above) gives valid and directly comparable values. To show this, outputs from a MeshGraphNets model were taken and RMSE was calculated between true and predicted values on the irregular grid and interpolated predictions. The RMSEs were with a few percent, indicating linearly interpolating predictions gives valid metrics.  

\section{Details on Baselines} \label{appendix:Baseline}
\textbf{MeshGraphNets.} For the MeshGraphNet baseline, we used the following parameters: 15 GNN message passing layers with hidden dimension 128, batch size 2, training horizon 5, noise magnitude 2e-2, pressure weight 0.1 and learning rate 1e-4 with an exponential decay schedule to 1e-6. These are identical to the original parameters proposed by \cite{MGN}, except we reduce the training horizon from 6 to 5 to reduce VRAM consumption. We tried manual hyperparameter tuning but did not find significantly better parameters. The training duration was 500 epochs. 

\textbf{DilResNet.} The DilResNet baseline closely follows the original implementation. There are 4 blocks of dilated convolutions each containing 7 CNN layers with 48 channels, 3x3 kernels, and ReLU activations. We trained with an initial learning rate of 1e-4 with an exponential decay schedule and the ADAM \cite{ADAM} optimizer along with training noise with magnitude 1e-3. 

\textbf{GAT2Conv.} This baseline is largely based on our decoder $f_\text{GNN}$. Similar to MeshGraphNets, this model operates directly on the irregular mesh. First, a MLP encoder encodes each node's values ($V_x, V_y, P$) which are then passed to a GNN with GAT2Conv layers which output predicted differences. Manual hyperparameter tuning was performed to optimize the GNN for standalone use. We found increasing the number of GNN heads from 1 to 4, layers from 3 to 10 and hidden size from 48 to 128 improved performance, while we kept the same learning rate and schedule. The purpose of this baseline is to ablate if a GAT2Conv decoder could be responsible for our model's accuracy, while the LLM doesn't do anything. The model was trained for 180 epochs, the same as the FLUID-OPT125m. 

The implementation of MeshGraphNets and DilResNet are loosely based on a PyTorch implementation from \cite{Eagle}, with some additional tuning and fixes.

\begin{figure} 
    \includegraphics[width=\textwidth]{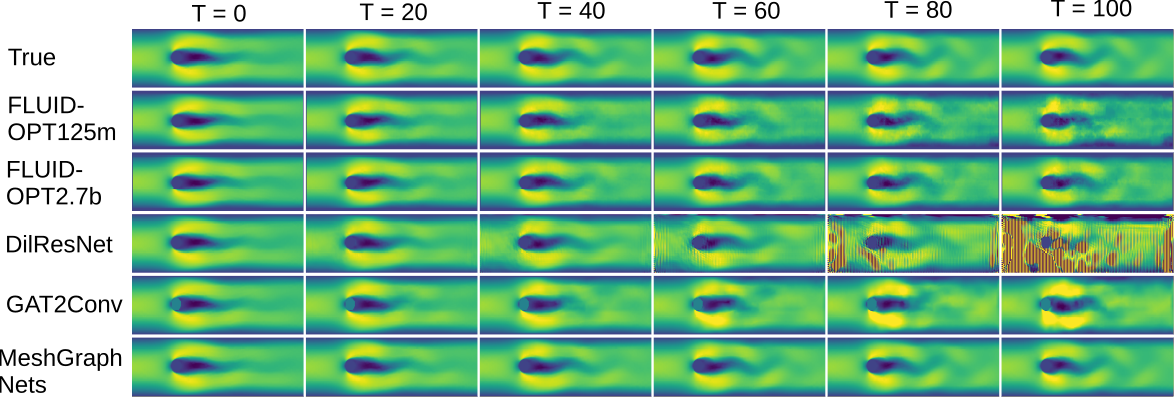}
    \caption{Plots of predictions for our models and baselines on the Cylinder dataset for the $V_x$ component, at different numbers of prediction steps. This particular sample involves simulating vortex instabilities for an incompressible fluid flow in a tube. }
    \label{fig:CylinderPlots}
\end{figure}

\begin{figure} 
    \includegraphics[width=\textwidth]{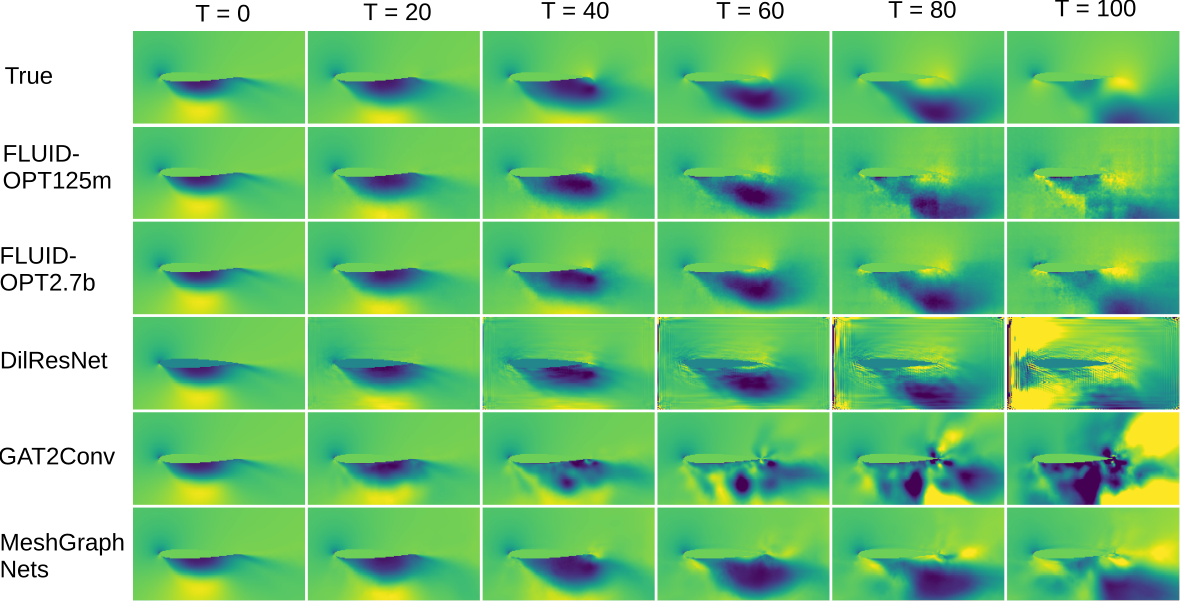}
    \caption{Plots of predictions for our models and baselines on the Airfoil dataset for the $V_x$ component, at different numbers of prediction steps.}
    \label{fig:AirfoilPlots}
\end{figure}

\section{Synthetic wave evolution dataset generation}  \label{Appendix:Synthetic}

\begin{figure} 
    \includegraphics[width=\textwidth]{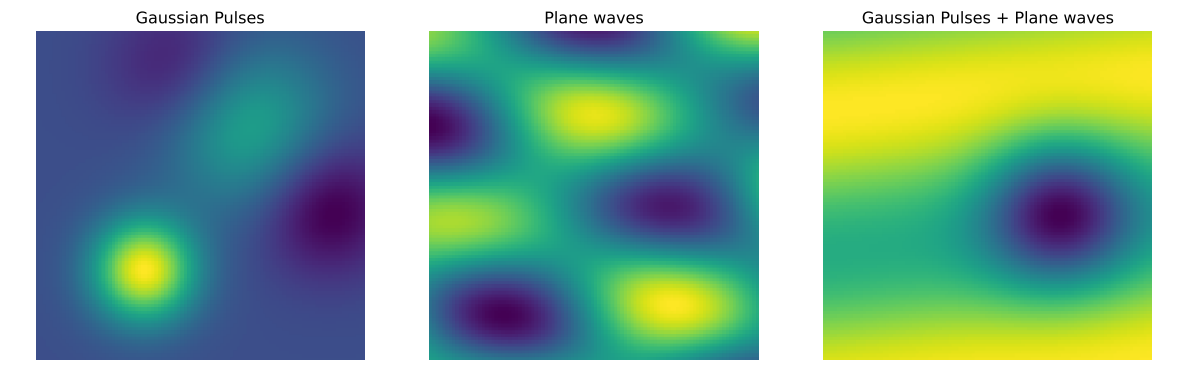}
    \caption{Examples of randomly initial conditions with Gaussian pulses, superimposed plane waves, and both Gaussian pulse and plane wave.}
    \label{fig:InitCond}
\end{figure}

\begin{figure} 
    \includegraphics[width=\textwidth]{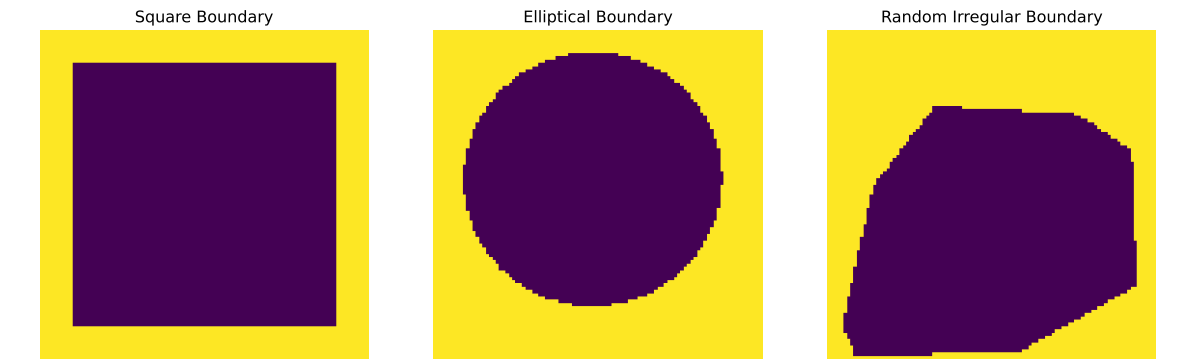}
    \caption{Examples of randomly generated square, elliptical, and irregular boundary conditions.}
    \label{fig:Bounary}
\end{figure}

\begin{figure} 
    \includegraphics[width=\textwidth]{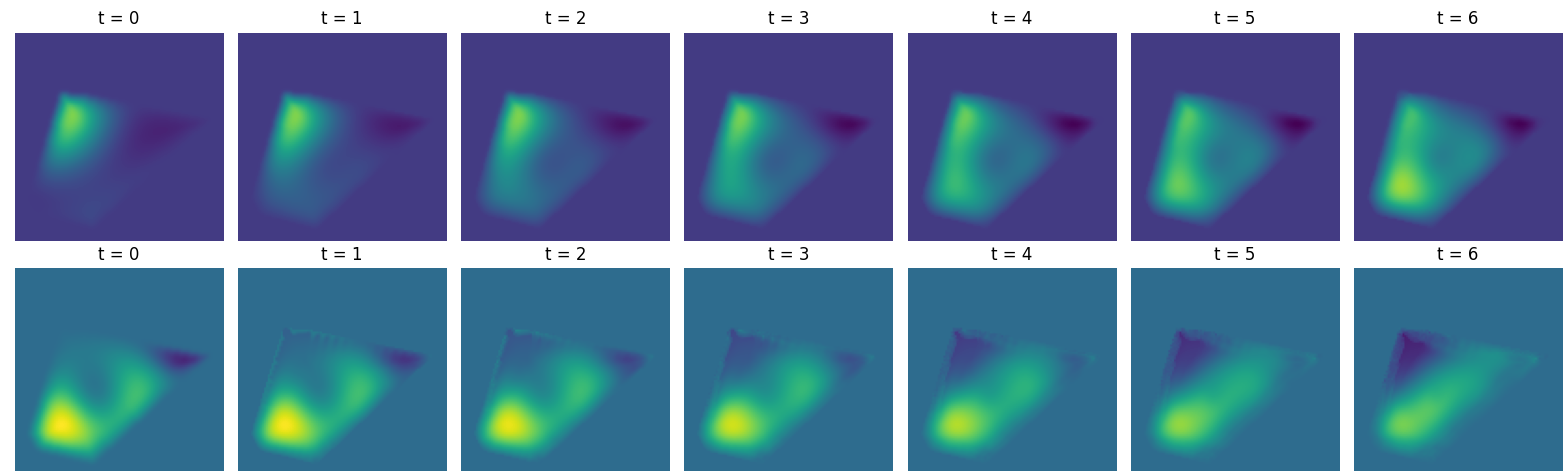}
    \caption{Samples from the synthetic dataset, with magnitudes $u(\textbf{x}, t)$ on the top and velocities $\frac{du(\textbf{x}, t)}{dt}$ on the bottom.}
    \label{fig:synth_rollout}
\end{figure}

The synthetic dataset is generated using Equation \ref{eq:synth}. As well as randomly selecting $c$, we randomly generate diverse initial and boundary conditions to encourage the model to learn the underlying equations for a wide range of conditions. 

Initial conditions are randomly chosen from 3 classes, 1) random Gaussian pulses, where a random number of Gaussian pulses of random size, amplitude, and position are superimposed, 2) random plane waves, where sinusoidal waves of random direction, phase, and wavelength are superimposed and 3) random waves and pulses, where a random plane wave and Gaussian pulse are superimposed. Examples of the 3 types are shown in Figure \ref{fig:InitCond}

Boundary conditions are set by applying boundary conditions $u(\delta x, t) = 0, \frac{du(\delta x, t)}{dt} = 0$ along a closed boundary $\delta x$. The shape of the boundary is also randomly generated from 3 classes, 1) rectangular boundary of random size, 2) elliptical boundary of random size and orientation, and 3) a random convex hull, generated by scattering 20 points in the domain and creating a convex hull from the points. Figure \ref{fig:Bounary} shows examples of generated boundary domains. 

Generating the initial state $s_0$ from the boundary and initial conditions is done by setting values outside the boundary to 0, multiplying the values inside the boundary by a smoothing factor proportional to the distance from the boundary, and smoothing using a Gaussian blur filter to ensure the initial state is smooth to avoid numerical instabilities. Both initial positions, $u(x, 0)$ and velocities $\frac{du(x, 0)}{dt}$ are randomly generated using this procedure. Starting from these initial conditions, Equation \ref{eq:synth} is integrated with the Torchdiffeq numerical ODE solver \cite{NODE_solver} using a numerical approximation of the Laplacian $\nabla^2$ and integrating through time. The states are recorded every $\Delta t = 0.05$, with 5 internal solver steps corresponding to $\Delta t = 0.01$ per step, with a grid size of $100 \times 100$ pixels. Finally, all states are normalized using dataset mean and variance statistics. An example of a simulation is shown in Figure \ref{fig:synth_rollout}

Since the equations are second order, the inputs to the model are the magnitudes $u(\textbf{x}, t)$ and velocities $\frac{du(\textbf{x}, t)}{dt}$ which are concatenated to form states $s_t = [u(\textbf{x}, t) || u(\textbf{x}, t) || \frac{du(\textbf{x}, t)}{dt}] \in \mathbb{R}^{N \times 3}$ for each timestep. Note we duplicate the magnitude since our model expects 3 input dimensions and we want to keep the model as similar to the other experiments as possible. Our model is trained on these states as described in the methodology section. 

In our experiments, we fix the random seed so trajectories are always identical and start predictions from $t=5$, with between 1 and 5 initial context states. The context must be kept fairly short, otherwise, information about $c$ leaks into every state and the model can deduce $c$ without using any context. This is because as the simulation progresses, different values of $c$ give different types of trajectories. 

\begin{figure} 
    \includegraphics[width=\textwidth]{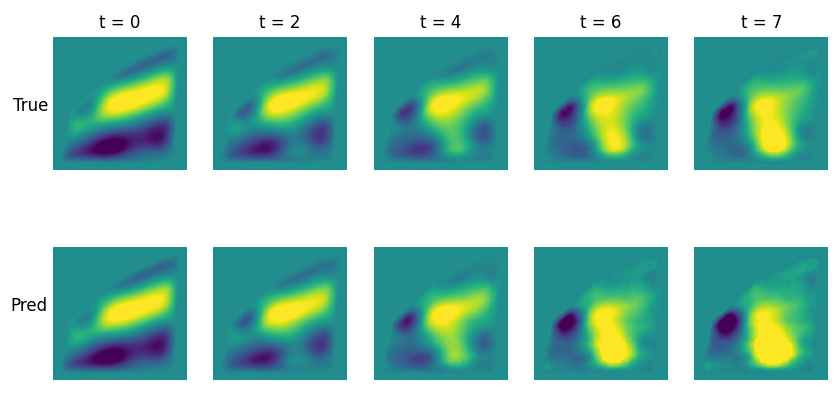}
    \caption{Plots of samples (top) and corresponding model predictions (bottom) of magnitude $u$ from the synthetic dataset at various steps from the first predicted states. }
    \label{fig:synth_preds}
\end{figure}

\end{document}